\documentclass{article}
\usepackage{ijcai16}
\usepackage{times}
\usepackage{multirow}
\usepackage{latexsym}
\usepackage{mathrsfs}
\usepackage{graphicx}
\usepackage{subfig}
\usepackage{amsfonts}
\usepackage{amssymb}
\usepackage{amsmath}
\usepackage{bm}
\usepackage{url}
\usepackage{pgfplots}

\usepackage{xcolor}
\usetikzlibrary{positioning,bayesnet}
\allowdisplaybreaks

\newenvironment{itemize*}%
  {\begin{itemize}%
    \setlength{\itemsep}{0pt}%
    \setlength{\parskip}{0pt}}%
  {\end{itemize}}
  \newenvironment{enumerate*}%
  {\begin{enumerate}%
    \setlength{\itemsep}{0pt}%
    \setlength{\parskip}{0pt}}%
  {\end{enumerate}}

\pgfplotsset{compat=1.5}

\DeclareMathOperator*{\softmax}{softmax}

\newcommand\newcite[1]{\citeauthor{#1} [\citeyear{#1}]}

\def\W{\mathbf{W}}
\def\U{\mathbf{U}}
\def\bb{\mathbf{b}}
\def\y{\mathbf{y}}
\def\V{\mathbf{V}}

\def\h{\mathbf{h}}

\def\x{\mathbf{x}}
\def\cc{\mathbf{c}}
\def\ii{\mathbf{i}}
\def\ff{\mathbf{f}}
\def\oo{\mathbf{o}}

\def\g{\mathbf{g}}

\title{Recurrent Neural Network for Text Classification \\with Multi-Task Learning}

\author{Pengfei Liu \quad Xipeng Qiu\thanks{Corresponding author.} \quad Xuanjing Huang\\
Shanghai Key Laboratory of Intelligent Information Processing, Fudan University\\
School of Computer Science, Fudan University\\
825 Zhangheng Road, Shanghai, China\\
\{pfliu14,xpqiu,xjhuang\}@fudan.edu.cn}

\date{}

\begin{document}
\maketitle
\begin{abstract}
Neural network based methods have obtained great progress on a variety of natural language processing tasks. However, in most previous works, the models are learned based on single-task supervised objectives, which often suffer from insufficient training data. In this paper, we use the multi-task learning framework to jointly learn across multiple related tasks. Based on recurrent neural network, we propose three different mechanisms of sharing information to model text with task-specific and shared layers. The entire network is trained jointly on all these tasks. Experiments on four benchmark text classification tasks show that our proposed models can improve the performance of a task with the help of other related tasks.
\end{abstract}

\section{Introduction}

Distributed representations of words  have been widely used in many natural language processing (NLP) tasks. Following this success, it is rising a substantial interest to learn the distributed representations of the continuous words, such as phrases, sentences, paragraphs and documents \cite{socher2013recursive,le2014distributed,kalchbrenner2014convolutional,liu2015multitimescale}. The primary role of these models is to represent the variable-length sentence or document as a fixed-length vector. A good representation of the variable-length text should fully capture the semantics of natural language.

The deep neural networks (DNN) based methods usually need a large-scale corpus due to the large number of parameters, it is hard to train a network that generalizes well with limited data.  However, the costs are extremely expensive to build the large scale resources for some NLP tasks. To deal with this problem, these models often involve an unsupervised pre-training phase.  The final model is fine-tuned with respect to a supervised training criterion with a gradient based optimization.
Recent studies have demonstrated significant accuracy gains in several NLP tasks \cite{collobert2011natural} with the help of the word representations learned from the large unannotated corpora.
Most pre-training methods are based on unsupervised objectives such as word prediction for training \cite{collobert2011natural,turian2010word,mikolov2013efficient}. This unsupervised pre-training is effective to improve the final performance, but it does not directly optimize the desired task.

Multi-task learning utilizes the correlation between related tasks to improve classification by learning tasks in parallel. Motivated by the success of multi-task learning \cite{caruana1997multitask}, there are several neural network based NLP models \cite{collobert2008unified,liu2015representation} utilize multi-task learning to jointly learn several tasks with the aim of mutual benefit.  The basic multi-task architectures of these models are to share some lower layers to determine common features. After the shared layers, the remaining layers are split into the multiple specific tasks.

In this paper, we propose three different models of sharing information with recurrent neural network (RNN). All the related tasks are integrated into a single system which is trained jointly. The first model uses just one shared layer for all the tasks. The second model uses different layers for different tasks, but each layer can read information from other layers. The third model not only assigns one specific layer for each task, but also builds a shared layer for all the tasks.
Besides, we introduce a gating mechanism to enable the model to selectively utilize the shared information. The entire network is trained jointly on all these tasks.

Experimental results on four text classification tasks show that the joint learning of multiple related tasks together can improve the performance of each task relative to learning them separately.

Our contributions are of two-folds:
\begin{itemize}
  \item First, we propose three multi-task architectures for RNN. Although the idea of multi-task learning is not new, our work is novel to integrate RNN into the multi-learning framework, which learns to map arbitrary text into semantic vector representations with both task-specific and shared layers.
  \item Second, we demonstrate strong results on several text classification tasks. Our multi-task models outperform most of state-of-the-art baselines.
\end{itemize}



\section{Recurrent Neural Network for Specific-Task Text Classification}

The primary role of the neural models is to represent the variable-length text as a fixed-length vector. These models generally consist of a projection layer that maps words, sub-word units or n-grams to vector representations (often trained beforehand with unsupervised methods), and then combine them with the different architectures of neural networks.

There are several kinds of models to model text, such as Neural Bag-of-Words (NBOW) model, recurrent neural network (RNN) \cite{chung2014empirical}, recursive neural network (RecNN) \cite{socher2012semantic,socher2013recursive} and convolutional neural network (CNN) \cite{collobert2011natural,kalchbrenner2014convolutional}. These models take as input the embeddings of words in the text sequence, and summarize its meaning with a fixed length vectorial representation.


Among them, recurrent neural networks (RNN) are one of the most popular architectures used in NLP problems because their recurrent structure is very suitable to process the variable-length text.

\subsection{Recurrent Neural Network}

A recurrent neural network (RNN) \cite{Elman:1990} is able to process a sequence of arbitrary length by recursively applying a
transition function to its \emph{internal hidden state vector} $h_t$ of the input sequence. The activation
 of the hidden state $h_t$ at time-step $t$ is computed as a function $f$ of the current input symbol
  $\x_t$ and the previous hidden state $\h_{t-1}$
\begin{equation}
    \h_t=
   \begin{cases}
   0 &\mbox{$t=0$}\\
   f(\h_{t-1},\x_t) &\mbox{otherwise}
   \end{cases}
\end{equation}

It is common to use the state-to-state transition function $f$ as the composition of an element-wise nonlinearity with an affine transformation of both $\x_t$ and $\h_{t-1}$.

Traditionally, a simple strategy for modeling sequence is to map the input sequence to a fixed-sized vector using one RNN, and then to feed the vector to a softmax layer for classification or other tasks \cite{cho2014learning}.

Unfortunately, a problem with RNNs with transition functions of this form is that during training, components of the gradient vector can grow or decay exponentially over long sequences \cite{hochreiter2001gradient,hochreiter1997long}.
This problem with \emph{exploding} or \emph{vanishing gradients} makes it difficult for the RNN model to learn long-distance correlations in a sequence.



Long short-term memory network (LSTM) was proposed by \cite{hochreiter1997long} to specifically address this issue of learning long-term dependencies. The LSTM maintains a separate memory cell inside it that updates and exposes its content only when deemed necessary.
A number of minor modifications to the standard LSTM unit have been made. While there are numerous LSTM variants, here we describe the implementation used by \newcite{graves2013generating}.

We define the LSTM \emph{units} at each time step $t$ to be a collection of vectors in $\mathbb{R}^d$: an \emph{input gate} $\ii_t$, a \emph{forget gate} $\ff_t$,  an \emph{output gate} $\oo_t$, a \emph{memory cell} $\cc_t$ and a hidden state $\h_t$. $d$ is the number of the LSTM units. The entries of the gating vectors $\ii_t$, $\ff_t$ and $\oo_t$ are in $[0, 1]$.
The LSTM transition equations are the following:
\begin{align}
\ii_t &=\sigma(\W_i\x_t+\U_i\h_{t-1}+\V_i\cc_{t-1}), \label{eq:lstm-eqs-1} \\
\ff_t &=\sigma(\W_f\x_t+\U_f\h_{t-1}+\V_f\cc_{t-1}), \label{eq:lstm-eqs-2}  \\
 \oo_t &=\sigma(\W_o\x_t+\U_o\h_{t-1}+\V_o\cc_t),  \label{eq:lstm-eqs-3} \\
 \tilde{\cc}{_t} &=\tanh(\W_c\x_t+\U_c\h_{t-1}), \label{eq:lstm-eqs-4} \\
\cc_t &=\ff_t^i \odot \cc_{t-1} + \ii_t \odot \tilde{\cc}{_t}, \label{eq:lstm-eqs-5} \\
\h_t &=\oo_t \odot \tanh(\cc_t) \label{eq:lstm-eqs-6},
\end{align}
where $x_t$ is the input at the current time step, $\sigma$ denotes the logistic sigmoid function and $\odot$ denotes elementwise multiplication. Intuitively, the forget gate controls the amount of which each unit of the memory cell is erased, the input gate controls how much each unit is updated, and the output gate controls the exposure of the internal memory state.

\subsection{Task-Specific Output Layer}
\begin{figure}[t]\centering
  \includegraphics[width=0.98\linewidth]{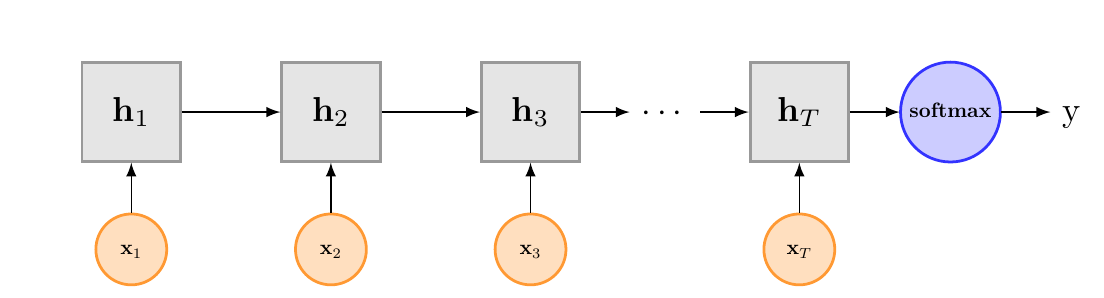}
  \caption{Recurrent Neural Network for Classification}\label{fig:rnn}
\end{figure}

In a single specific task, a simple strategy is to map the input sequence to a fixed-sized vector using one RNN, and then to feed the vector to a softmax layer for classification or other tasks.


Given a text sequence $x = \{x_1, x_2, \cdots, x_T\}$, we first use a lookup layer to get the vector representation (embeddings) $\x_i$ of the each word $x_i$.
The output at the last moment $\h_T$  can be regarded as the representation of the whole sequence, which has a fully connected layer followed by a softmax non-linear layer that predicts the probability distribution over classes.

Figure \ref{fig:rnn} shows the unfolded RNN structure for text classification.

The parameters of the network are trained to minimise the cross-entropy of the predicted and true distributions.

\begin{equation}
  L( {\hat{\y}}, \y) = - \sum_{i=1}^{N} \sum_{j=1}^C  \y_i^j \log(\hat{\y}_i^j),
\end{equation}
where $\y_i^j$ is the ground-truth label; ${\hat{y}_i^j}$ is prediction probabilities; $N$ denotes the number of training samples and $C$ is the class number.  

\section{Three Sharing Models for RNN based Multi-Task Learning}

Most existing neural network methods are based on supervised training objectives on a single task \cite{collobert2011natural,socher2013recursive,kalchbrenner2014convolutional}. These methods often suffer from the limited amounts of training data. To deal with this problem, these models often involve an unsupervised pre-training phase. This unsupervised pre-training is effective to improve the final performance, but it does not directly optimize the desired task.

Motivated by the success of multi-task learning \cite{caruana1997multitask}, we propose three multi-task models to leverage supervised data from many related tasks. Deep neural model is well suited for multi-task learning since the features learned from a task may be useful for other tasks. Figure \ref{fig:ours} gives an illustration of our proposed models.

{\begin{figure}[t]
  \centering \hspace{-3em}
  \subfloat[Model-I: Uniform-Layer Architecture]{
  \includegraphics[width=0.5\textwidth]{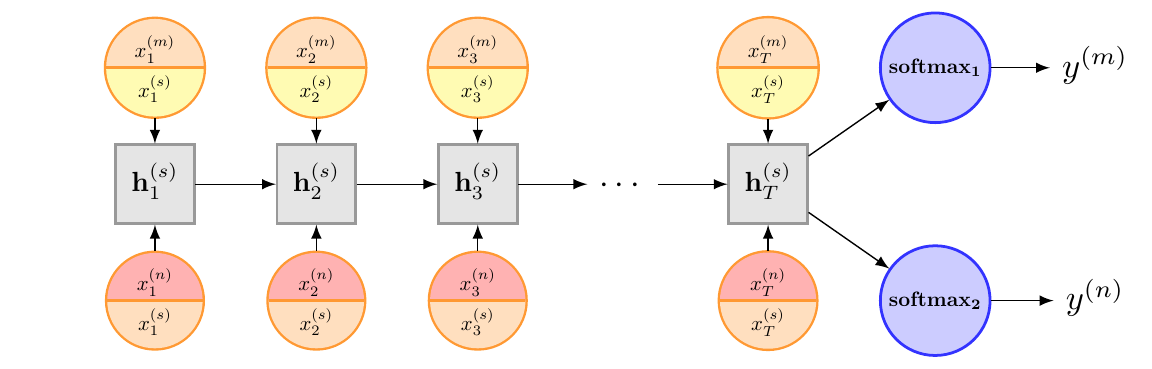}
  }\\  \hspace{-3em}
  \subfloat[Model-II: Coupled-Layer Architecture]{
  \includegraphics[width=0.5\textwidth]{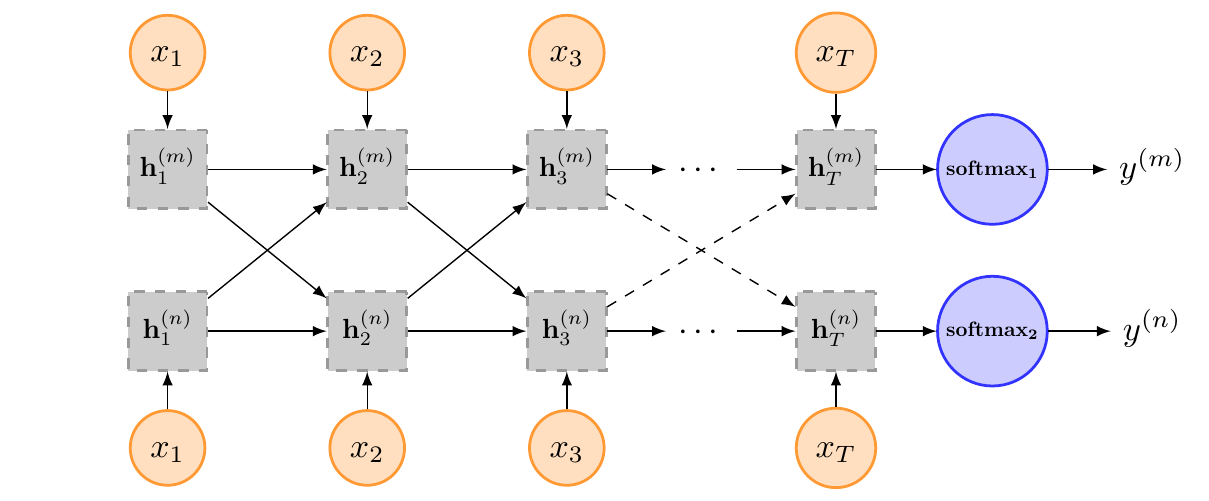}
  }\\\hspace{-3em}
  \subfloat[Model-III: Shared-Layer Architecture]{
  \includegraphics[width=0.5\textwidth]{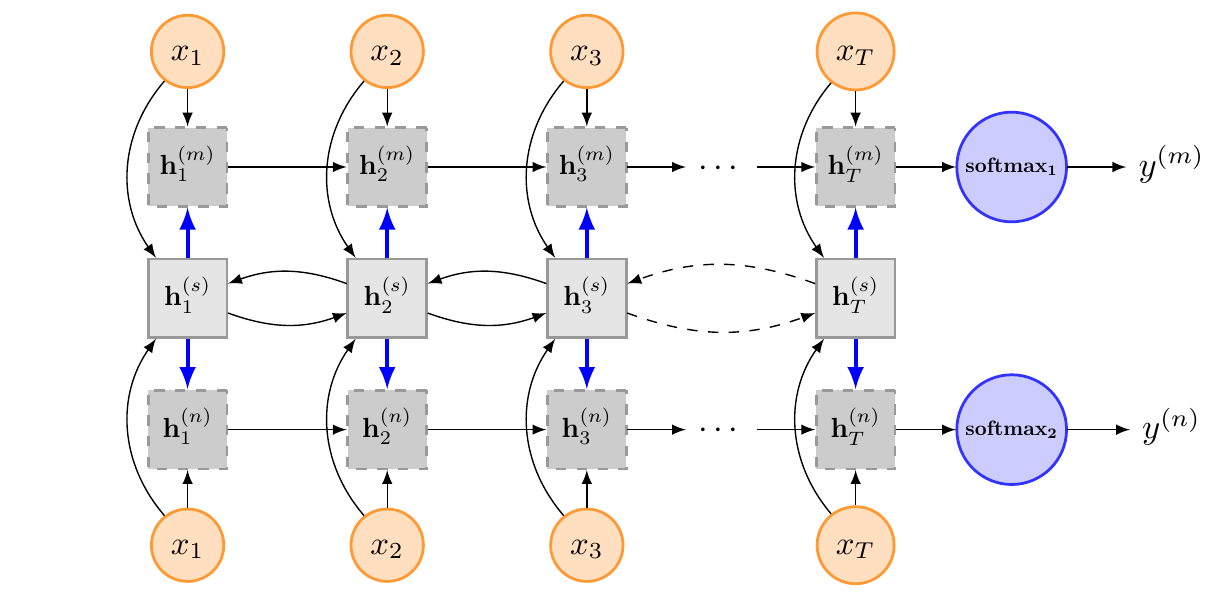}
  }
  \caption{Three architectures for modelling text with multi-task learning.
   }\label{fig:ours}
\end{figure}
}

\paragraph{Model-I: Uniform-Layer Architecture}
In Model-I, the different tasks share a same LSTM layer and an embedding layer besides their own embedding layers.

For task $m$, the input $\hat{\x}_t$ consists of two parts:
\begin{align}
\hat{\x}_t^{(m)} = \x_t^{(m)} \oplus \x_t^{(s)},
\end{align}
where $\x_t^{(m)}$ , $\x_t^{(s)}$ denote the task-specific and shared word embeddings respectively, $\oplus$ denotes the concatenation operation.

The LSTM layer is shared for all tasks. The final sequence representation for task $m$ is
the output of LSMT at step $T$.
 \begin{align}
\h_{T}^{(m)} = LSTM(\hat{\x}^{(m)}).\label{eq:task_emb}
\end{align}


\paragraph{Model-II: Coupled-Layer Architecture}

In Model-II, we assign a LSTM layer for each task, which can use the information for the LSTM layer of the other task.

Given a pair of tasks $(m,n)$, each task has own LSTM in the task-specific model. We denote the outputs at step $t$ of two coupled LSTM layer are $\h_{t}^{(m)}$ and $\h_{t}^{(n)}$.

To better control signals flowing from one task to another task, we use a global gating unit which endows the model with the capability of deciding how much information it should accept. We re-define Eqs. (\ref{eq:lstm-eqs-4}) and the new memory content of an LSTM at $m$-th task is computed by:
\begin{equation}\small
 \tilde{\cc}{_t}^{(m)} =\tanh\left(\W_c^{(m)}\x_t + \sum_{i\in \{m,n\}}\g^{(i\rightarrow m)}U_c^{(i\rightarrow m)} \h_{t-1}^{(i)} \right)
\end{equation}
where $\g^{(i\rightarrow m)} = \sigma(\W_g^{(m)} \x_t + \U_g^{(i)} \h_{t-1}^{(i)})$. The other settings are same to the standard LSTM.

This model can be used to jointly learning for every two tasks. We can get two task specific representations $\h_T^{(m)}$ and $\h_T^{(n)}$ for tasks $m$ and $n$ receptively.

\begin{table*}[!th]\centering
  \begin{tabular}{|*{8}{c|}}
    \hline
    Dataset & Type & Train Size & Dev. Size& Test Size & Class & Averaged Length & Vocabulary Size  \\
    \hline
    SST-1 & Sentence & 8544 & 1101 & 2210 & 5 & 19 & 18K\\
    SST-2 & Sentence & 6920 & 872 & 1821 & 2 & 18 & 15K\\

    SUBJ & Sentence & 9000 & - & 1000 & 2 & 21 & 21K\\
    IMDB & Document & 25,000 & - & 25,000 & 2 & 294 & 392K\\
    \hline
  \end{tabular}
  \caption{Statistics of the four datasets used in this paper.}\label{tab:data}
\end{table*}

\paragraph{Model-III: Shared-Layer Architecture}

Model-III also assigns a separate LSTM layer for each task, but introduces a bidirectional LSTM layer to capture the shared information for all the tasks.

We denote the outputs of the forward and backward LSTMs at step $t$ as $\overrightarrow \h_{t}^{(s)}$ and  $\overleftarrow \h_{t}^{(s)}$ respectively. The output of shared layer is $\h_t^{(s)} = \overrightarrow \h_{t}^{(s)} \oplus \overleftarrow \h_{t}^{(s)}$.

To enhance the interaction between task-specific layers and the shared layer, we use gating mechanism to endow the neurons in task-specific layer with the ability to accept or refuse the information passed by the neuron in shared layers.
Unlike Model-II, we compute the new state for LSTM as follows:
\begin{equation}\small
 \tilde{\cc}_t^{(m)} =\tanh\left(\W_c^{(m)} \x_t + \g^{(m)} \U_c^{(m)}  \h_{t-1}^{(m)} + \g^{(s\rightarrow m)} \U_c^{(s)} \h_t^{(s)} \right),
\end{equation}
where $\g^{(m)} = \sigma(\W_g^{(m)} \x_t + \U_g^{(m)} \h_{t-1}^{(m)})$ and $\g^{(s\rightarrow m)} = \sigma(\W_g^{(m)} \x_t + \U_g^{(s\rightarrow m)} \h_t^{(s)})$.

\section{Training}

The  task-specific representations, which emittd by the muti-task architectures of all of the above, are ultimately fed into different output layers, which are also task-specific.
\begin{align}
{\hat{\y}}^{(m)} = \softmax(\W^{(m)}\h^{(m)} + \bb^{(m)}),
\end{align}
where ${\hat{\y}}^{(m)}$ is prediction probabilities for task $m$, $\W^{(m)}$ is the weight which needs to be learned, and $\bb^{(m)}$ is a bias term.

Our global cost function is the linear combination of cost function for all joints.
\begin{align}
\phi = \sum_{m=1}^{M}{\lambda}_m  L({\hat{y}}^{(m)}, y^{(m)})
\end{align}
where $\lambda_m$ is the weights for each task $m$ respectively.

It is worth noticing that labeled data for training each task can come from completely different datasets. Following \cite{collobert2008unified}, the training is achieved in a stochastic manner by looping over the tasks:
\begin{enumerate}
  \item Select a random task.
  \item Select a random training example from this task.
  \item Update the parameters for this task by taking a gradient step with respect to this example.
  \item Go to 1.
\end{enumerate}

\vspace{-1em}
\paragraph{Fine Tuning}
For model-I and model-III, there is a shared layer for all the tasks. Thus, after the joint learning phase, we can use a fine tuning strategy to further optimize the performance for each task.

\vspace{-1em}
\paragraph{Pre-training of the shared layer with neural language model}
For model-III, the shared layer can be initialized by an unsupervised pre-training phase. Here, for the shared LSTM layer in Model-III, we initialize it by a language model \cite{bengio2007greedy}, which is trained on all the four task dataset.

\section{Experiment}

In this section, we investigate the empirical performances of our proposed three models on four related text classification tasks and then compare it to other state-of-the-art models.

\subsection{Datasets}

To show the effectiveness of multi-task learning, we choose four different text classification tasks about movie review.
Each task have own dataset, which is briefly described as follows.

\begin{itemize}
  \item \textbf{SST-1} The movie reviews with five classes (negative, somewhat negative, neutral, somewhat positive, positive) in the Stanford Sentiment Treebank\footnote{\url{http://nlp.stanford.edu/sentiment}.} \cite{socher2013recursive}.
  \item \textbf{SST-2} The movie reviews with binary classes. It is also from the Stanford Sentiment Treebank.
  \item \textbf{SUBJ} Subjectivity data set where the goal is to classify each instance (snippet) as being subjective or objective. \cite{pang2004sentimental}
  \item \textbf{IMDB} The IMDB dataset\footnote{\url{http://ai.stanford.edu/~amaas/data/sentiment/}} consists of 100,000 movie reviews with binary classes  \cite{maas2011learning}. One key aspect of this dataset is that each movie review has several sentences. 
\end{itemize}

The first three datasets are sentence-level, and the last dataset is document-level. The detailed statistics about the four datasets are listed in Table \ref{tab:data}.

\subsection{Hyperparameters and Training}
The network is trained with backpropagation and the gradient-based optimization is performed using the Adagrad update rule \cite{duchi2011adaptive}.
In all of our experiments, the word embeddings are trained using word2vec \cite{mikolov2013efficient} on the Wikipedia corpus (1B words). The vocabulary size is about 500,000. The word embeddings are fine-tuned during training to improve the performance \cite{collobert2011natural}. The other parameters are initialized by randomly sampling from uniform distribution in [-0.1, 0.1].
The hyperparameters which achieve the best performance on the development set will be chosen for the final evaluation. For datasets without development set, we use 10-fold cross-validation (CV) instead.

The final hyper-parameters are as follows. The embedding size for specific task and shared layer are 64. For Model-I, there are two embeddings for each word, and both their sizes are 64. The hidden layer size of LSTM is 50. The initial learning rate is $0.1$. The regularization weight of the parameters is $10^{-5}$.

\subsection{Effect of Multi-task Training}

\begin{table}[!th]\small
\center
\begin{tabular}{|l|*{6}{c|}}
\hline
\textbf{Model} &	 SST-1 &	 SST-2 &	SUBJ&	 IMDB  & Avg$\Delta$\\
\hline
Single Task &   45.9    &   85.8    &   91.6    &   88.5    & - \\
\hline
Joint Learning &   46.5    &   86.7    &   92.0    &   89.9  & +0.8   \\
+ Fine Tuning &   \textbf{48.5}    &   \textbf{87.1}    &   \textbf{93.4}    &   \textbf{90.8} & +2.0   \\
\hline
\end{tabular}
\caption{Results of the uniform-layer architecture. }\label{tab:result-1}
\end{table}

\begin{table}[!th]\small
\center
\begin{tabular}{|l|*{6}{c|}}
\hline
\textbf{Model} &	 SST-1 &	 SST-2 &	SUBJ&	 IMDB  & Avg$\Delta$\\
\hline
Single Task &   45.9    &   85.8    &   91.6    &   88.5    & -\\
\hline
SST1-SST2 &   \textbf{48.9}    &   \textbf{87.4}    &   -    &   -    & +2.3\\
SST1-SUBJ &   46.3    &   -    &   92.2    &   -    & +0.5\\
SST1-IMDB &   46.9    &   -    &   -    &   89.5    & +1.0\\
SST2-SUBJ &   -    &   86.5    &   92.5    &   -    & +0.8\\
SST2-IMDB &   -    &   86.8    &   -    &   \textbf{89.8}    & +1.2\\
SUBJ-IMDB &   -    &   -    &   \textbf{92.7}    &   89.3    & +0.9\\
\hline
\end{tabular}
\caption{Results of the coupled-layer architecture.}\label{tab:result-2}
\end{table}

\begin{table}[!th]\small
\center
\begin{tabular}{|l|*{6}{c|}}
\hline
\textbf{Model} &	 SST-1 &	 SST-2 &	SUBJ&	 IMDB & Avg$\Delta$\\
\hline
Single Task &   45.9    &   85.8    &   91.6    &   88.5   & - \\
\hline
Joint Learning &   47.1    &   87.0    &   92.5    &   90.7 & +1.4   \\
+ LM &   47.9    &   86.8    &   93.6    &   91.0   & +1.9 \\
+ Fine Tuning &   \textbf{49.6}    &   \textbf{87.9}    &   \textbf{94.1}    &   \textbf{91.3}   & +2.8 \\
\hline
\end{tabular}
\caption{Results of the shared-layer architecture.}\label{tab:result-3}
\end{table}

We first compare our our proposed models with the standard LSTM for single task classification. We use the implementation
of \newcite{graves2013generating}. The unfolded illustration is shown in Figure \ref{fig:rnn}.

Table \ref{tab:result-1}-\ref{tab:result-3} show the classification accuracies on the four datasets. The second line (``Single Task'') of each table shows the result of the standard LSTM for each individual task.

\paragraph{Uniform-layer Architecture}
For the first uniform-layer architecture, we train the model on four datasets simultaneously. The LSTM layer is shared across all the tasks. The average improvement of the performances on four datasets is 0.8\%. With the further fine-tuning phase, the improvement achieves 2.0\% on average.

\paragraph{Coupled-layer Architecture}
For the second coupled-layer architecture, the information is shared with a pair of tasks. Therefore, there are six combinations for the four datasets. We train six models on the different pairs of datasets. We can find that the pair-wise joint learning also improves the performances. The more relevant the tasks are, the more significant the improvements are. Since SST-1 and SST-2 are from the same corpus, their improvements are more significant than the other combinations. The improvement is 2.3\% on average with simultaneously learning on SST-1 and SST-2.

\paragraph{Shared-layer Architecture}

The shared-layer architecture is more general than uniform-layer architecture. Besides a shared layer for all the tasks, each task has own task-specific layer. As shown in Table \ref{tab:result-3}, we can see that the average improvement of the performances on four datasets is 1.4\%, which is better than the uniform-layer architecture. We also investigate the strategy of unsupervised pre-training towards shared LSTM layer. With the LM pre-training, the performance is improved by an extra 0.5\% on average. Besides, the further fine-tuning can significantly improve the performance by an another 0.9\%.

To recap, all our proposed models outperform the baseline of single-task learning. The shared-layer architecture gives the best performances. Moreover, compared with  vanilla LSTM, our proposed three models don't cause much extra computational cost while  converge faster. In our experiment, the most complicated model-III, costs  2.5 times as long as vanilla LSTM.

\subsection{Comparisons with State-of-the-art Neural Models}

\begin{table}[!th]\small
\center
\begin{tabular}{|l|*{5}{c|}}
\hline
\textbf{Model} &	 SST-1 &	 SST-2 &	SUBJ&	 IMDB \\

\hline
NBOW & 42.4 & 80.5 & 91.3 & 83.62 \\
MV-RNN  &	 44.4 &	 82.9 &	 -&	 - \\
RNTN  &	 45.7 &	 85.4 &	 -&	 - \\
DCNN  &	 48.5&	 86.8 &	 - &	 - \\
PV  &	 44.6&	 82.7 & 90.5 &	 \textbf{91.7} \\
Tree-LSTM  &	 \textbf{50.6}&	 86.9 & - & - \\
\hline
Multi-Task &   49.6    &   \textbf{87.9}    &   \textbf{94.1}    &    91.3    \\

\hline
\end{tabular}
\caption{Results of shared-layer multi-task model against state-of-the-art neural models.}\label{tab:result-4}
\end{table}

\begin{figure*}[!t]
\centering
   \subfloat[]{
   \includegraphics[width=0.49\textwidth,height=7em]{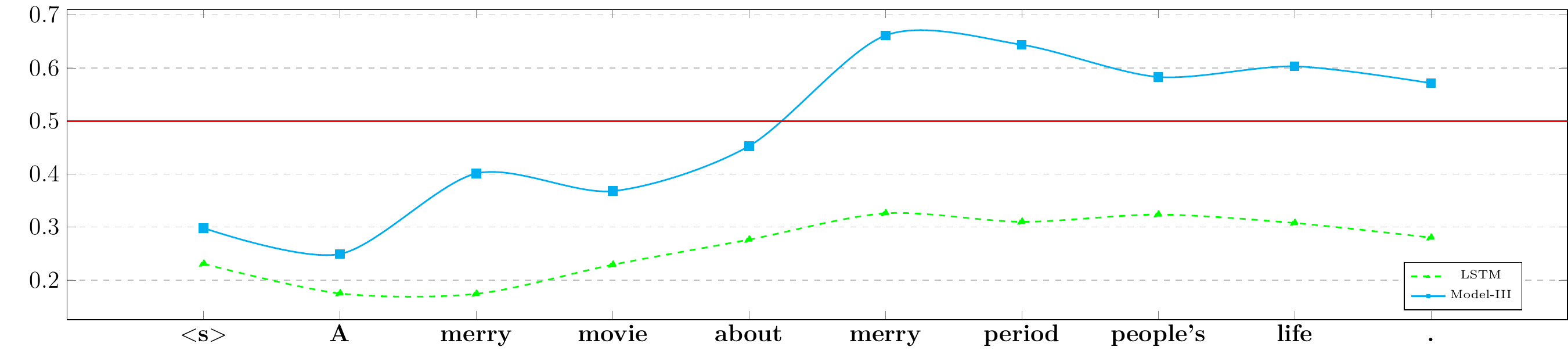}
   }
   \subfloat[]{
   \includegraphics[width=0.49\textwidth,height=7em]{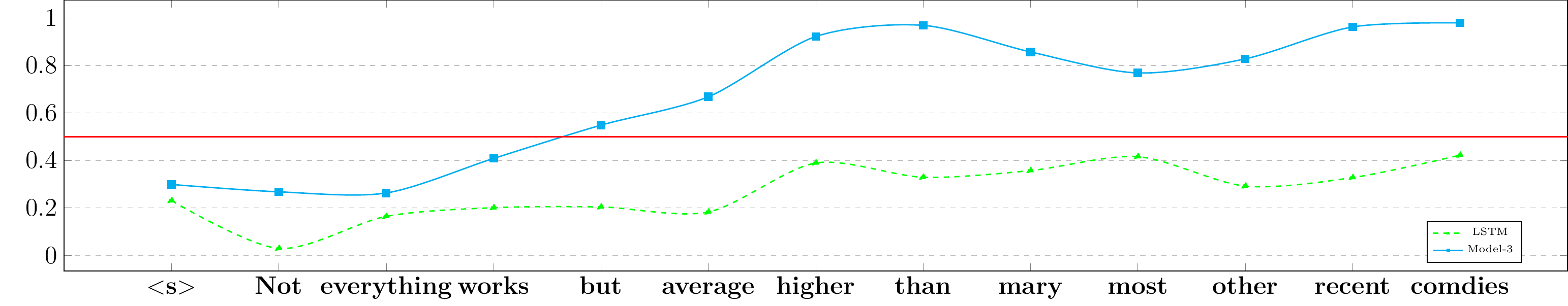}
   }\\
   \subfloat[]{
   \includegraphics[width=0.49\textwidth,height=7em]{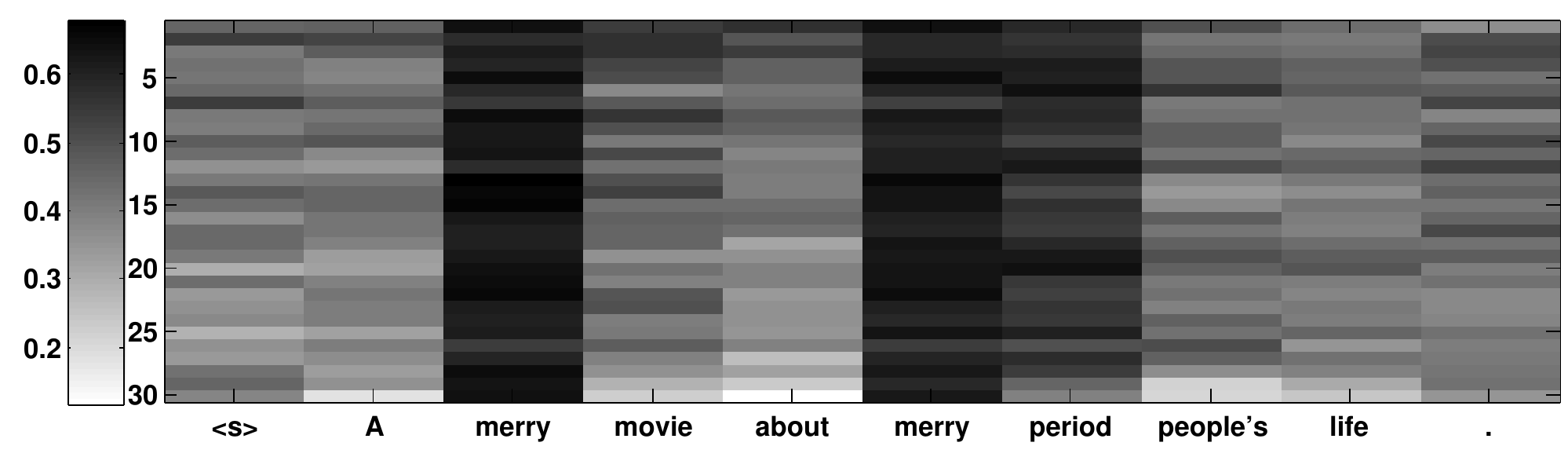}
   }
   \subfloat[]{
   \includegraphics[width=0.49\textwidth,height=7em]{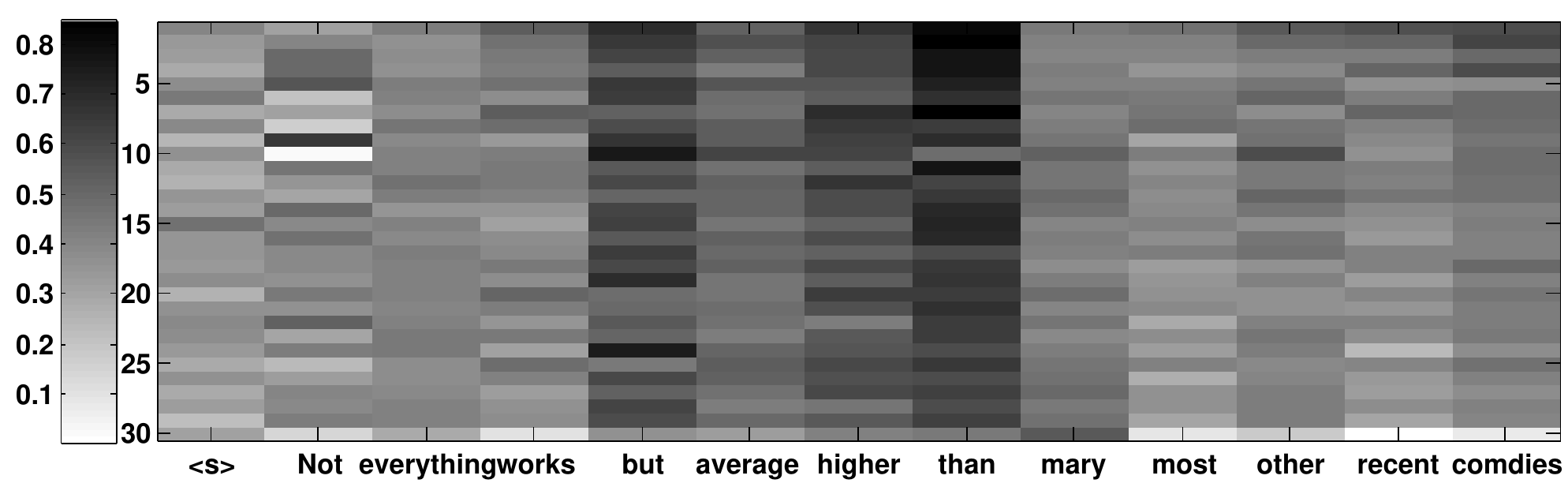}
   }
  \caption{(a)(b) The change of the predicted sentiment score at different time steps. Y-axis represents the sentiment score, while X-axis represents the input words in chronological order. The red horizontal line gives a border between the positive and negative sentiments. (c)(d) Visualization of the global gate's ($g^{(s)}$) activation.
}\label{fig:exp-case}
\end{figure*}

We compare our model with the following models:

\begin{itemize}
\item \textbf{NBOW} The NBOW sums the word vectors and applies a non-linearity followed by a softmax classification layer. 
\item  \textbf{MV-RNN} Matrix-Vector Recursive Neural Network with parse trees \cite{socher2012semantic}.
\item \textbf{RNTN} Recursive Neural Tensor Network with tensor-based feature function and parse trees \cite{socher2013recursive}.
\item \textbf{DCNN} Dynamic Convolutional Neural Network with dynamic k-max pooling \cite{kalchbrenner2014convolutional}.
\item \textbf{PV} Logistic regression on top of paragraph vectors \cite{le2014distributed}. Here, we use the popular open source implementation of PV in Gensim\footnote{\url{https://github.com/piskvorky/gensim/}}.
\item \textbf{Tree-LSTM} A generalization of LSTMs to tree-structured network topologies. \cite{tai2015improved}
\end{itemize}

Table \ref{tab:result-4} shows the performance of the shared-layer architecture compared with the competitor models, which shows our model is competitive for the neural-based state-of-the-art models.

Although Tree-LSTM outperforms our model on SST-1, it needs an external parser to get the sentence topological structure. It is worth noticing that our models are compatible with the other RNN based models. For example, we can easily extend our models to incorporate the Tree-LSTM model.

\subsection{Case Study}

To get an intuitive understanding of what is happening when we use the single LSTM or the shared-layer LSTM to
predict the class of text, we design an experiment to analyze the output of the single LSTM and the shared-layer LSTM at each time step.
We sample two sentences from the SST-2 test dataset, and the changes of the predicted sentiment score at different time steps are shown in Figure \ref{fig:exp-case}. To get more insights into how the shared structures influences the specific task.
We observe the activation of global gates $g^{(s)}$, which controls signals flowing from one shared LSTM layer to task-spcific layer, to understand the behaviour of neurons.
We plot evolving activation of global gates $g^{(s)}$ through time and sort the neurons according to their activations at the last time step.

For the sentence ``\texttt{A merry movie about merry period people's life.}'', which has a positive sentiment, while the standard LSTM gives a wrong prediction. The reason can be inferred from the activation of global gates $g^{(s)}$.  As shown in Figure \ref{fig:exp-case}-(c), we can see clearly the neurons are activated much when they take input as ``\texttt{merry}'', which indicates the task-specific layer takes much information from shared layer towards the word ``\texttt{merry}'', and this ultimately makes the model give a correct prediction.

Another case ``\texttt{Not everything works, but the average is higher than in Mary and most other recent comedies.}''  is positive and has a little complicated semantic composition. As shown in Figure \ref{fig:exp-case}-(b,d), simple LSTM cannot capture the structure of ``\texttt{but ... higher than }'' while our model is sensitive to it, which indicates the shared layer can not only enrich the meaning of certain words, but can teach some information of structure to specific task.

\subsection{Error Analysis}
We analyze the bad cases induced by our proposed shared-layer model on SST-2 dataset. Most of the bad cases can be generalized into
two categories
\vspace{-1em}
\paragraph{Complicated Sentence Structure}
Some sentences involved complicated structure can not be handled properly, such as double negation ``\texttt{it never fails to engage us.}'' and  subjunctive sentences ``\texttt{Still, I thought it could have been more.}''. To solve these cases, some architectural improvements are necessary, such as tree-based LSTM \cite{tai2015improved}.

\vspace{-1em}
\paragraph{Sentences Required Reasoning}
The sentiments of some sentences can be mislead if only considering the literal meaning. For example, the sentence ``\texttt{I tried to read the time on my watch.}'' expresses  negative attitude towards a movie, which can be understood correctly by reasoning based on common sense.

\section{Related Work}

Neural networks based multi-task learning has proven effective in many NLP problems \cite{collobert2008unified,liu2015representation}. 

\newcite{collobert2008unified} used a shared representation for input words and solve different traditional NLP tasks such as part-of-Speech tagging and semantic role labeling within one framework. However, only one lookup table is shared, and the other lookup-tables and layers are task specific.
To deal with the variable-length text sequence, they used window-based method to fix the input size.

\newcite{liu2015representation} developed a multi-task DNN for learning representations across multiple tasks. Their multi-task DNN approach combines tasks of query classification and ranking for web search. But the input of the model is bag-of-word representation, which lose the information of word order.

Different with the two above methods, our models are based on recurrent neural network, which is better to model the  variable-length text sequence.

More recently, several multi-task encoder-decoder networks were also proposed for  neural machine translation \cite{dong2015multi,firat2016multi}, which can make use of cross-lingual information. Unlike these works, in this paper we design three architectures, which can control the information flow between shared layer and task-specific layer flexibly, thus obtaining better sentence representations.

\section{Conclusion and Future Work}
In this paper, we introduce three RNN based architectures to model text sequence with multi-task learning. The differences among them are the mechanisms of sharing information among the several tasks.
Experimental results show that our models can improve the performances of a group of related tasks by exploring common features.

In future work, we would like to investigate the other sharing mechanisms of the different tasks.

\section*{Acknowledgments}
We would like to thank the anonymous reviewers for their valuable comments. This work was partially funded by National Natural Science Foundation of China (No. 61532011, 61473092, and 61472088), the National High Technology Research and Development Program of China (No. 2015AA015408).


\begin{thebibliography}{}

\bibitem[\protect\citeauthoryear{Bengio \bgroup \em et al.\egroup
  }{2007}]{bengio2007greedy}
Yoshua Bengio, Pascal Lamblin, Dan Popovici, Hugo Larochelle, et~al.
\newblock Greedy layer-wise training of deep networks.
\newblock {\em Advances in neural information processing systems}, 19:153,
  2007.

\bibitem[\protect\citeauthoryear{Caruana}{1997}]{caruana1997multitask}
Rich Caruana.
\newblock Multitask learning.
\newblock {\em Machine learning}, 28(1):41--75, 1997.

\bibitem[\protect\citeauthoryear{Cho \bgroup \em et al.\egroup
  }{2014}]{cho2014learning}
Kyunghyun Cho, Bart van Merrienboer, Caglar Gulcehre, Fethi Bougares, Holger
  Schwenk, and Yoshua Bengio.
\newblock Learning phrase representations using rnn encoder-decoder for
  statistical machine translation.
\newblock In {\em Proceedings of EMNLP}, 2014.

\bibitem[\protect\citeauthoryear{Chung \bgroup \em et al.\egroup
  }{2014}]{chung2014empirical}
Junyoung Chung, Caglar Gulcehre, KyungHyun Cho, and Yoshua Bengio.
\newblock Empirical evaluation of gated recurrent neural networks on sequence
  modeling.
\newblock {\em arXiv preprint arXiv:1412.3555}, 2014.

\bibitem[\protect\citeauthoryear{Collobert and
  Weston}{2008}]{collobert2008unified}
Ronan Collobert and Jason Weston.
\newblock A unified architecture for natural language processing: Deep neural
  networks with multitask learning.
\newblock In {\em Proceedings of ICML}, 2008.

\bibitem[\protect\citeauthoryear{Collobert \bgroup \em et al.\egroup
  }{2011}]{collobert2011natural}
Ronan Collobert, Jason Weston, L{\'e}on Bottou, Michael Karlen, Koray
  Kavukcuoglu, and Pavel Kuksa.
\newblock Natural language processing (almost) from scratch.
\newblock {\em The Journal of Machine Learning Research}, 12:2493--2537, 2011.

\bibitem[\protect\citeauthoryear{Dong \bgroup \em et al.\egroup
  }{2015}]{dong2015multi}
Daxiang Dong, Hua Wu, Wei He, Dianhai Yu, and Haifeng Wang.
\newblock Multi-task learning for multiple language translation.
\newblock In {\em Proceedings of the ACL}, 2015.

\bibitem[\protect\citeauthoryear{Duchi \bgroup \em et al.\egroup
  }{2011}]{duchi2011adaptive}
John Duchi, Elad Hazan, and Yoram Singer.
\newblock Adaptive subgradient methods for online learning and stochastic
  optimization.
\newblock {\em The Journal of Machine Learning Research}, 12:2121--2159, 2011.

\bibitem[\protect\citeauthoryear{Elman}{1990}]{Elman:1990}
Jeffrey~L Elman.
\newblock Finding structure in time.
\newblock {\em Cognitive science}, 14(2):179--211, 1990.

\bibitem[\protect\citeauthoryear{Firat \bgroup \em et al.\egroup
  }{2016}]{firat2016multi}
Orhan Firat, Kyunghyun Cho, and Yoshua Bengio.
\newblock Multi-way, multilingual neural machine translation with a shared
  attention mechanism.
\newblock {\em arXiv preprint arXiv:1601.01073}, 2016.

\bibitem[\protect\citeauthoryear{Graves}{2013}]{graves2013generating}
Alex Graves.
\newblock Generating sequences with recurrent neural networks.
\newblock {\em arXiv preprint arXiv:1308.0850}, 2013.

\bibitem[\protect\citeauthoryear{Hochreiter and
  Schmidhuber}{1997}]{hochreiter1997long}
Sepp Hochreiter and J{\"u}rgen Schmidhuber.
\newblock Long short-term memory.
\newblock {\em Neural computation}, 9(8):1735--1780, 1997.

\bibitem[\protect\citeauthoryear{Hochreiter \bgroup \em et al.\egroup
  }{2001}]{hochreiter2001gradient}
Sepp Hochreiter, Yoshua Bengio, Paolo Frasconi, and J{\"u}rgen Schmidhuber.
\newblock Gradient flow in recurrent nets: the difficulty of learning long-term
  dependencies, 2001.

\bibitem[\protect\citeauthoryear{Kalchbrenner \bgroup \em et al.\egroup
  }{2014}]{kalchbrenner2014convolutional}
Nal Kalchbrenner, Edward Grefenstette, and Phil Blunsom.
\newblock A convolutional neural network for modelling sentences.
\newblock In {\em Proceedings of ACL}, 2014.

\bibitem[\protect\citeauthoryear{Le and Mikolov}{2014}]{le2014distributed}
Quoc~V. Le and Tomas Mikolov.
\newblock Distributed representations of sentences and documents.
\newblock In {\em Proceedings of ICML}, 2014.

\bibitem[\protect\citeauthoryear{Liu \bgroup \em et al.\egroup
  }{2015a}]{liu2015multitimescale}
PengFei Liu, Xipeng Qiu, Xinchi Chen, Shiyu Wu, and Xuanjing Huang.
\newblock Multi-timescale long short-term memory neural network for modelling
  sentences and documents.
\newblock In {\em Proceedings of the Conference on Empirical Methods in Natural
  Language Processing}, 2015.

\bibitem[\protect\citeauthoryear{Liu \bgroup \em et al.\egroup
  }{2015b}]{liu2015representation}
Xiaodong Liu, Jianfeng Gao, Xiaodong He, Li~Deng, Kevin Duh, and Ye-Yi Wang.
\newblock Representation learning using multi-task deep neural networks for
  semantic classification and information retrieval.
\newblock In {\em NAACL}, 2015.

\bibitem[\protect\citeauthoryear{Maas \bgroup \em et al.\egroup
  }{2011}]{maas2011learning}
Andrew~L Maas, Raymond~E Daly, Peter~T Pham, Dan Huang, Andrew~Y Ng, and
  Christopher Potts.
\newblock Learning word vectors for sentiment analysis.
\newblock In {\em Proceedings of the ACL}, pages 142--150, 2011.

\bibitem[\protect\citeauthoryear{Mikolov \bgroup \em et al.\egroup
  }{2013}]{mikolov2013efficient}
Tomas Mikolov, Kai Chen, Greg Corrado, and Jeffrey Dean.
\newblock Efficient estimation of word representations in vector space.
\newblock {\em arXiv preprint arXiv:1301.3781}, 2013.

\bibitem[\protect\citeauthoryear{Pang and Lee}{2004}]{pang2004sentimental}
Bo~Pang and Lillian Lee.
\newblock A sentimental education: Sentiment analysis using subjectivity
  summarization based on minimum cuts.
\newblock In {\em Proceedings of ACL}, 2004.

\bibitem[\protect\citeauthoryear{Socher \bgroup \em et al.\egroup
  }{2011}]{socher2011semi}
Richard Socher, Jeffrey Pennington, Eric~H Huang, Andrew~Y Ng, and
  Christopher~D Manning.
\newblock Semi-supervised recursive autoencoders for predicting sentiment
  distributions.
\newblock In {\em Proceedings of EMNLP}, 2011.

\bibitem[\protect\citeauthoryear{Socher \bgroup \em et al.\egroup
  }{2012}]{socher2012semantic}
Richard Socher, Brody Huval, Christopher~D Manning, and Andrew~Y Ng.
\newblock Semantic compositionality through recursive matrix-vector spaces.
\newblock In {\em Proceedings of EMNLP}, pages 1201--1211, 2012.

\bibitem[\protect\citeauthoryear{Socher \bgroup \em et al.\egroup
  }{2013}]{socher2013recursive}
Richard Socher, Alex Perelygin, Jean~Y Wu, Jason Chuang, Christopher~D Manning,
  Andrew~Y Ng, and Christopher Potts.
\newblock Recursive deep models for semantic compositionality over a sentiment
  treebank.
\newblock In {\em Proceedings of EMNLP}, 2013.

\bibitem[\protect\citeauthoryear{Tai \bgroup \em et al.\egroup
  }{2015}]{tai2015improved}
Kai~Sheng Tai, Richard Socher, and Christopher~D Manning.
\newblock Improved semantic representations from tree-structured long
  short-term memory networks.
\newblock {\em arXiv preprint arXiv:1503.00075}, 2015.

\bibitem[\protect\citeauthoryear{Turian \bgroup \em et al.\egroup
  }{2010}]{turian2010word}
Joseph Turian, Lev Ratinov, and Yoshua Bengio.
\newblock Word representations: a simple and general method for semi-supervised
  learning.
\newblock In {\em Proceedings of ACL}, 2010.

\end{thebibliography}

\end{document}